\documentclass{article}

\usepackage[final]{neurips_2024}

\usepackage{amsmath,amsfonts,bm}

\def\eqref#1{equation~\ref{#1}}

\def\1{\bm{1}}

\DeclareMathAlphabet{\mathsfit}{\encodingdefault}{\sfdefault}{m}{sl}
\SetMathAlphabet{\mathsfit}{bold}{\encodingdefault}{\sfdefault}{bx}{n}

\usepackage{authblk}
\usepackage{hyperref}

\usepackage{url}
\usepackage{booktabs}
\usepackage{tikz}
\usepackage{wrapfig}
\usepackage{graphicx}
\usepackage{subcaption}
\usepackage{caption}
\usetikzlibrary{matrix,decorations.pathreplacing,arrows.meta,positioning, calc, shapes.geometric, backgrounds}
\definecolor{lavender}{rgb}{0.96, 0.73, 1.0}
\definecolor{brickred}{rgb}{0.8, 0.25, 0.33}
\newcommand{\std}[1]{\textcolor{gray}{\(\pm\) \footnotesize{#1}}}
\definecolor{mygreen}{RGB}{0, 153, 0}

\title{On the Efficiency of NLP-Inspired Methods for Tabular Deep Learning}

\author[1]{\textbf{\small Anton Frederik Thielmann} \thanks{Correspondence to \texttt{anton.thielmann@basf.com}}}
\author[2]{\textbf{\small Soheila Samiee}}

\affil[1]{\small BASF SE, Germany}
\affil[2]{\small BASF Canada Inc., Canada}
\affil[ ]{\{firstname.lastname\}@basf.com}

\begin{document}
\maketitle

\begin{abstract}
Recent advancements in tabular deep learning (DL) have led to substantial performance improvements, surpassing the capabilities of traditional models.
With the adoption of techniques from natural language processing (NLP), such as language model-based approaches, DL models for tabular data have also grown in complexity and size. 
Although tabular datasets do not typically pose scalability issues, the escalating size of these models has raised efficiency concerns. Despite its importance, efficiency has been relatively underexplored in tabular DL research. This paper critically examines the latest innovations in tabular DL, with a dual focus on performance and computational efficiency.
The source code is available at \url{https://github.com/basf/mamba-tabular}.
\end{abstract}

\section{Introduction}

Deep Neural Networks have emerged as a powerful tool for a wide range of tasks, particularly those involving unstructured data. In areas such as image classification \citep{mauricio2023comparing}, video processing \citep{apostolidis2021video}, point cloud analysis \citep{zhang2023deep}, protein structure prediction \citep{baek2021accurate, jumper2021highly}, or natural language understanding (e.g. \citep{brown2020language}), the best-performing models today are all based on neural networks. 

Particularly for textual problems, larger models consistently outperform their smaller counterparts \citep{muennighoff2022mteb}. Therefore, efficiency is of critical importance, not only for inference but also for training. As many models, especially large language models, rely on transformer architectures, the quadratic time complexity of attention mechanisms with respect to sequence length can be especially challenging for longer sequences \citep{beltagy2020longformer}.

Recently, the introduction of the Mamba architecture \citep{gu2023Mamba}, based on state-space models (SSMs) \citep{gu2021efficiently}, has gained significant attention due to its promise of a linear increase in memory consumption compared to the quadratic increase observed in transformer architectures. Mamba is designed to be both memory-efficient and high-performing, as demonstrated by the variety of extensions built upon it. For instance, the MoE (JAMBA) architecture \citep{lieber2024jamba, pioro2024moe}, Mamba for point clouds \citep{liu2024point, zhang2024point}, for time series \citep{ahamed2024timemachine, liang2024bi}, and for graphs \citep{behrouz2024graph} all show the versatility and efficiency of Mamba-based models.

Tabular data tasks, on the other hand, remain one of the last frontiers not yet fully dominated by deep neural networks. Gradient Boosted Decision Trees (GBDTs) still lead the way for tasks involving tabular data \citep{mcelfresh2024neural, grinsztajn2022tree}. However, tabular deep learning has started to close the performance gap. Advances from natural language processing have shown promise for tabular tasks. Models such as TabTransformer \citep{huang2020tabtransformer}, FT-Transformer \citep{gorishniy2021revisiting} and Mambular \citep{thielmann2024mambular} have proven to be competitive models, occasionally outperforming GBDTs. More recently, Prior-Fitted Networks (PFNs) have introduced a new and promising direction for tabular models \citep{hollmann2022tabpfn}. Despite this, models like TabPFN \citep{hollmann2022tabpfn} are not yet scalable for larger datasets. 

Since tabular tasks rarely involve massive datasets, efficiency has traditionally not been a significant concern. However, to truly close the gap between GBDTs and tabular deep learning models, both, memory-efficiency and speed will play a crucial role, since GBDTs are not only performant but also highly efficient compared to tabular deep learning models \citep{gu2021efficiently, nori2019interpretml}. While architectures like transformers and Mamba, originally designed for NLP, have been adapted for tabular tasks, a thorough analysis of their efficiency for tabular data is still missing. This paper addresses that gap by evaluating and comparing the performance and efficiency of several tabular deep learning models, complementing existing benchmark studies that focus solely on performance (e.g., \citep{mcelfresh2024neural, gorishniy2021revisiting, grinsztajn2022tree, borisov2022deep}).

In summary, the contributions of this paper are as follows: 
\begin{enumerate} 
    \renewcommand{\labelenumi}{\Roman{enumi}.} 
    \item We evaluate LLM-inspired deep learning methods for tabular data, focusing on their efficiency and examining how effectively efficient NLP techniques can be applied to tabular tasks
    \item We demonstrate the effectiveness of recurrent neural networks (RNNs) and show that simple sequential architectures can be powerful tools for solving tabular problems. 
\end{enumerate}

\section{Methodology}
We follow the experimental design from \cite{thielmann2024mambular}.
The exact same pre-processing, data splits and evaluation is used. The notation is aligned with \cite{thielmann2024mambular} and \cite{gorishniy2021revisiting} such that $\mathcal{D} = \{ (\bm{x}^{(i)}, y^{(i)})\}_{i=1}^n$ is the training dataset of size $n$ and $y$ denotes the target variable that can be arbitrarily distributed. Each input $\bm{x} = (x_1, x_2, \dots, x_J)$ contains $J$ features (variables).
The categorical and numerical features are denoted as $\bm{x} \equiv (\bm{x}_{cat}, \bm{x}_{num})$.
For each feature, embedding friendly piecewise linear encodings as presented by \cite{gorishniy2022embeddings} are used.

\begin{equation}\label{eq:ple}
z_{j(\text{num})}^{t} = \begin{cases} 
0 & \text{if } x < b_{t-1}, \\
1 & \text{if } x \geq b_t, \\
\frac{x - b_{t-1}}{b_{t} - b_{t-1}} & \text{else.}
\end{cases}
\end{equation}

The tested models can be grouped into Transformer-based models, such as FT-Transformer \citep{gorishniy2021revisiting}; Recurrent models, including Mambular \citep{thielmann2024mambular} and a classical RNN; hybrid architectures, similar to the Jamba architecture \citep{lieber2024jamba}; and MLP and ResNet architectures.

For all models except the MLP and ResNet, the features are fed through embedding layers with identical dimensions for all datasets. For categorical features, traditional embedding layers are used after integer encoding. For numerical features, simple linear layers with the same dimensions as the embedding layers are employed. After the embedding layer, in the transformer and recurrent models, the feature matrix has the shape $\tilde{\mathbf{x}} \in \mathbb{R}^{N \times J \times d}$, where $N$ is the batch size, $J$ is the number of variables, and $d$ is the embedding dimension. 

This paper introduces two new architectures for tabular deep learning that are described in the following:

\paragraph{TabulaRNN}
TabulaRNN is taking advantage of a RNN architecture. Instead of token embeddings, the feature embeddings, similar to those in Mambular \citep{thielmann2024mambular} are passed to the RNN.  
The RNN model processes an input sequence \(\mathbf{x} = [\mathbf{x}_1, \mathbf{x}_2, \dots, \mathbf{x}_T]\), where \(\mathbf{x}_t\) represents the input at time step \(t\). At each time step, the hidden state \(\mathbf{h}_t\) is updated based on the previous hidden state \(\mathbf{h}_{t-1}\) and the current input \(\mathbf{x}_t\). In this context, the \textit{sequence} the RNN iterates over corresponds to the features of the tabular data, where each feature represents a step in the sequence. The hidden state is computed using the following equation:
\begin{equation}
    \mathbf{h}_t = \sigma(\mathbf{W}_h \mathbf{h}_{t-1} + \mathbf{W}_x \tilde{\mathbf{x}}_t + \mathbf{b}),
\end{equation}

where \(\mathbf{W}_h\) describes recurrent weight matrix applied to the previous hidden state and \(\mathbf{W}_x\) is the input weight matrix applied to the current input. \(\mathbf{b}\) describes a simple bias term, and \(\sigma\) is an activation function.

The final output representation, which is passed to a tabular MLP head, is computed by averaging over the hidden states, i.e., over the hidden state for each variable. 
Since this corresponds to each sequence having a fixed length, missing values can be handled by introducing special tokens, using imputation methods such as median imputation, or simply dropping them. All model configuration details can be found here: \url{https://github.com/basf/mamba-tabular}.

\paragraph{MambAttention}
MambAttention is a combination of Mambular \citep{thielmann2024mambular} and the FT-Transformer \citep{gorishniy2021revisiting}. It loosely follows the modeling architecture from \cite{lieber2024jamba}. The input sequence is embedded just like in the RNN. The first layer is a Mamba Block, with the given matrices:
\[
\mathbf{A} \in \mathbb{R}^{1 \times 1 \times d \times \delta}, \quad
\mathbf{B} \in \mathbb{R}^{N \times J \times 1 \times \delta}, \quad
\Delta \in \mathbb{R}^{N \times J \times d \times 1}, \bar{\mathbf{z}} \in \mathbb{R}^{N \times J \times d \times 1}
,\]
where $\delta$ denotes a inner dimension, $\bar{\mathbf{z}}$ has the same entries as $\tilde{\textbf{x}}$, but one additional axis. The hidden states, \(\mathbf{h}_j \in \mathbb{R}^{N \times d\times \delta}\), are updated as follows: 
\begin{equation}\label{eq:ssm}
\mathbf{h}_j = \exp\left(\Delta \odot_3 \mathbf{A} \right)_{:, j, :, :}
\odot_{1,2,3} \mathbf{h}_{j-1} + \left (\left( \Delta \odot_{1,2} \mathbf{B} \right) \odot_{1,2,3} \bar{\mathbf{z}}\right )_{:, j, :, :}.
\end{equation}
After the first Mamba block follows a standard multihead-attention layer. The sequences state is simply passed from the Mamba block to the attention block. These blocks are designed in alternating fashion with the first and the last block always being Mamba blocks. The final representation is achieved by averaging over the hidden states. This representation is passed to a final MLP task head. 

\paragraph{Mambular-Triton} 
For evaluation of efficiency of Mambular model, in addition to a purely pytorch implemented Mambular, a Triton based Mambular version\footnote{For all Mambular-Triton experiments, Mamba1 from the mamba-ssm \href{https://github.com/state-spaces/mamba}{package}, version 2.2.2 is used.} -- closely following the original Mamba implementation \citep{gu2023Mamba} and \citep{thielmann2024mambular} -- is tested. This did not have any impact on performance, so the result of its implementation is only reported for effeciency. In the result section, the pure pytorch implementation is denoted as \textit{Mambular} and the version following the Mamba1 implementation is denoted as \textit{mambular-Triton}.

\section{Results} 

\subsection{Performance}
The performance of the models are benchmarked using 12 open-source data sets across two tasks (regression and classification). Details of datasets are provided in the supplementary materials \ref{app:datasets}. All experimental results were obtained using 5-fold cross-validation. The average results and standard deviations across the 5 folds are reported. All models utilize PLE encodings (Eq. \ref{eq:ple}). 
The results are reported 
in Table \ref{tab:performance_result}. Overall results confirm that sequence-based deep learning models perform very well for tabular tasks. Even the simple TabulaRNN shows strong performance, comparable to both Mambular and FT-Transformer. Alternating attention layers and Mamba blocks does not lead to any performance improvement. Overall, Mambular is the best performing model, closely followed by the FT-Transformer and TabulaRNN. 

\begin{table}[ht!]
\centering
\caption{Benchmarking results for the regression and classification tasks. Average mean squared error values and average AUC values over 5 folds and the corresponding standard deviations are reported, respectively. Smaller values are better for regression and larger values are better in classification tasks (marked with arrows). The best performing model is marked in bold. 
}
\resizebox{\textwidth}{!}{
\begin{tabular}{l|ccccccc|ccccc||c}
\toprule
  &  \multicolumn{7}{c}{Regression Tasks} & \multicolumn{5}{c}{Classification Tasks} & Avg. Rank\\
Models  & DI $\downarrow$ & AB $\downarrow$ & CA $\downarrow$ & WI $\downarrow$ & PA $\downarrow$ & HS $\downarrow$ & CP $\downarrow$ & BA $\uparrow$ & AD $\uparrow$ & CH $\uparrow$ & FI $\uparrow$ & MA $\uparrow$ &  \\
\midrule
FT-Transformer*  & 0.018 & 0.458 & 0.169 & \textbf{0.615} & \textbf{0.024} & \textbf{0.111} & \textbf{0.024} & 0.926 & 0.926 & \textbf{0.863} & 0.792 & 0.916 & 2.25\\
                 & \std{0.001} & \std{0.055} & \std{0.006} & \std{0.012} & \std{0.005} & \std{0.014} & \std{0.001} & \std{0.003} & \std{0.002} & \std{0.007} & \std{0.011} & \std{0.003} & \\
MLP*             & 0.066 & 0.462 & 0.198 & 0.654 & 0.764 & 0.147 & 0.031 & 0.895 & 0.914 & 0.840 & 0.793 & 0.886 & 5.58\\
                 & \std{0.003} & \std{0.051} & \std{0.011} & \std{0.013} & \std{0.023} & \std{0.017} & \std{0.001} & \std{0.004} & \std{0.002} & \std{0.005} & \std{0.011} & \std{0.003} & \\
ResNet*          & 0.039 & 0.455 & 0.178 & 0.639 & 0.606 & 0.141 & 0.030 & 0.896 & 0.917 & 0.841 & 0.793 & 0.889 & 4.25\\
                 & \std{0.018} & \std{0.045} & \std{0.006} & \std{0.013} & \std{0.031} & \std{0.017} & \std{0.002} & \std{0.006} & \std{0.002} & \std{0.006} & \std{0.013} & \std{0.003} & \\
Mambular*        & \textbf{0.018} & \textbf{0.452} & \textbf{0.167} & 0.628 & 0.035 & 0.132 & 0.027 & 0.927 & \textbf{0.928} & 0.856 & 0.795 & 0.917 & \textbf{2.00}\\
                 & \std{0.000} & \std{0.043} & \std{0.011} & \std{0.010} & \std{0.005} & \std{0.020} & \std{0.002} & \std{0.006} & \std{0.002} & \std{0.004} & \std{0.011} & \std{0.003} & \\
MambAttention         & \textbf{0.018} & 0.484 & 0.189 & 0.638 & 0.030 & 0.142 & 0.026 & 0.919 & 0.921 & 0.857 & 0.781 & 0.911 & 3.67\\
                 & \std{0.000} & \std{0.052} & \std{0.006} & \std{0.003} & \std{0.006} & \std{0.024} & \std{0.002} & \std{0.004} & \std{0.002} & \std{0.004} & \std{0.009} & \std{0.001} & \\
TabulaRNN        & \textbf{0.018} & 0.459 & 0.178 & 0.659 & 0.073 & 0.114 & 0.027 & \textbf{0.930} & 0.925 & 0.855 & \textbf{0.796} & \textbf{0.922} & 2.75\\
                 & \std{0.000} & \std{0.047} & \std{0.013} & \std{0.013} & \std{0.012} & \std{0.014} & \std{0.001} & \std{0.004} & \std{0.002} & \std{0.006} & \std{0.011} & \std{0.002} & \\
\bottomrule
  \multicolumn{3}{c}{* adopted from \cite{thielmann2024mambular}} & \\

\end{tabular}
}

\label{tab:performance_result}
\end{table}

\subsection{Efficiency}
The memory-efficiency and computation time of all six models have been benchmarked in a more controlled set up. 
Each model is designed with approximately 350k trainable parameters to ensure comparability in terms of size. All models are analyzed on the same Nvidia-T4 GPU. Simulated data, with an equal number of numerical and categorical features is used for all efficiency results. Where applicable, the models use an embedding size of 64 which is equal to each numerical feature dimension corresponding to 64 bins in PLE encodings. Categorical features were limited to 10 categories. The results are obtained with a simulated mini batch size of 32 for simulations with less than 100 features and 8 for simulations with more than 100 features.

\begin{figure}[hb!]
    \centering
     \includegraphics[width=0.45\textwidth]{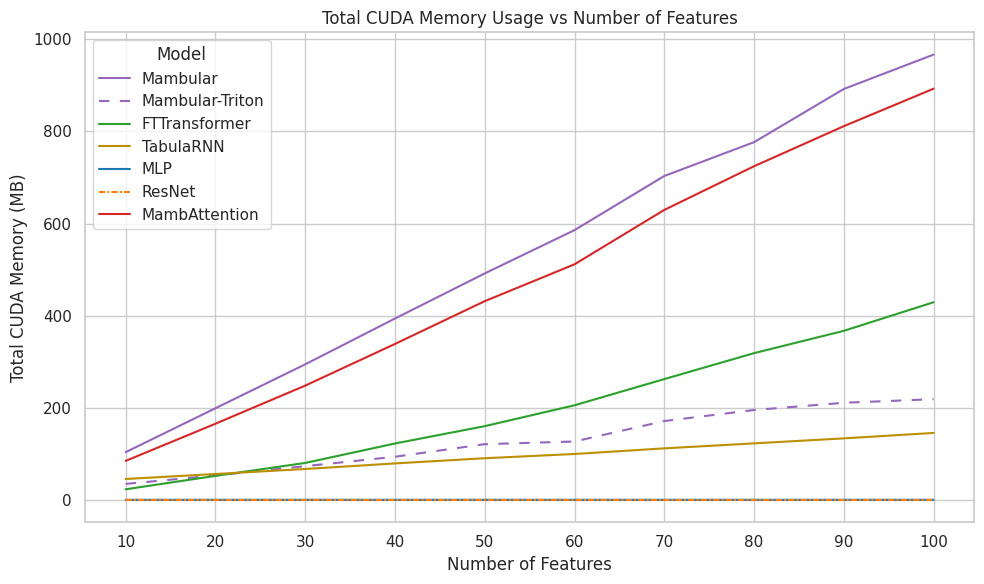}
     \includegraphics[width=0.45\textwidth]{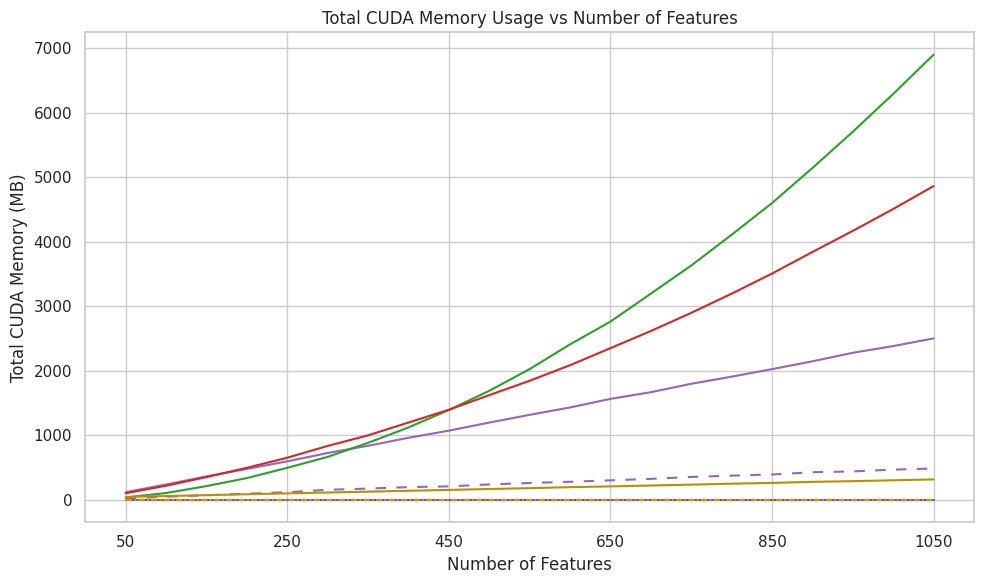}
    \caption{\footnotesize{Memory consumption as a function of the number of features. Left shows GPU memory usage for datasets with $<$100 features/variables. MambAttention and Mambular show the highest memory usage, increasing linearly with the number of features. Mambular-Triton's GPU memory consumption is significantly lower than base Mambular. (Right) For a large number of features, FT-Transformer's memory usage increases quadratically, approaching Mambular and MambAttention at around 400 features.}}
    \label{fig:memory_consumption_features}
\end{figure}

In terms of memory usage, both ResNet and MLP are significantly superior compared to other models, exhibiting the lowest GPU consumption across different feature sets. The GPU consumption against number of features is visualized in Figure \ref{fig:memory_consumption_features}. TabulaRNN, while less performant than Mambular and FT-Transformer, stands out for its high efficiency, showing linear GPU usage as the number of features increases. In contrast, FT-Transformer's GPU consumption grows quadratically, making it less efficient for larger feature sets. However, for problems with more than 25 features, the Triton version of Mambular becomes more efficient than the attention-based FT-Transformer. On the other hand, in the pure PyTorch version, FT-Transformer remains more efficient than Mambular for problems with fewer than 350 features in our specific configuration. Given, that tabular problems most often do not entail such large numbers of features -- the maximum number of features in Table~\ref{tab:performance_result} is 32 -- Triton Mambular implementation or the transformer based models would be efficient alternatives compared to the recurrent pytorch Mambular model.
MambAttention, on the other hand, proved to be neither efficient nor particularly performant, falling behind the other models in both respects. 

\begin{figure}[ht!]
    \centering
    \begin{minipage}[b]{0.48\textwidth}
        \centering
        \includegraphics[width=\linewidth]{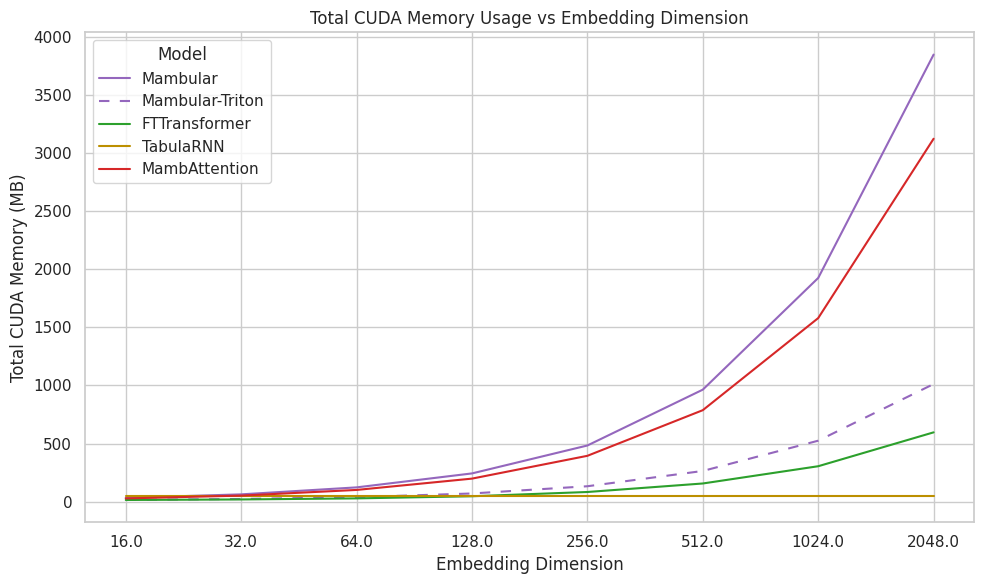}
        \caption{\footnotesize{Memory consumption as a function of embedding dimension for a fixed set of 12 features. A pure pytorch Mambular's memory consumption increases significantly faster than that of the FT-Transformer.}}
        \label{fig:embedding_dimension}
    \end{minipage}%
    \hfill
    \begin{minipage}[b]{0.48\textwidth}
        \centering
        \includegraphics[width=\linewidth]{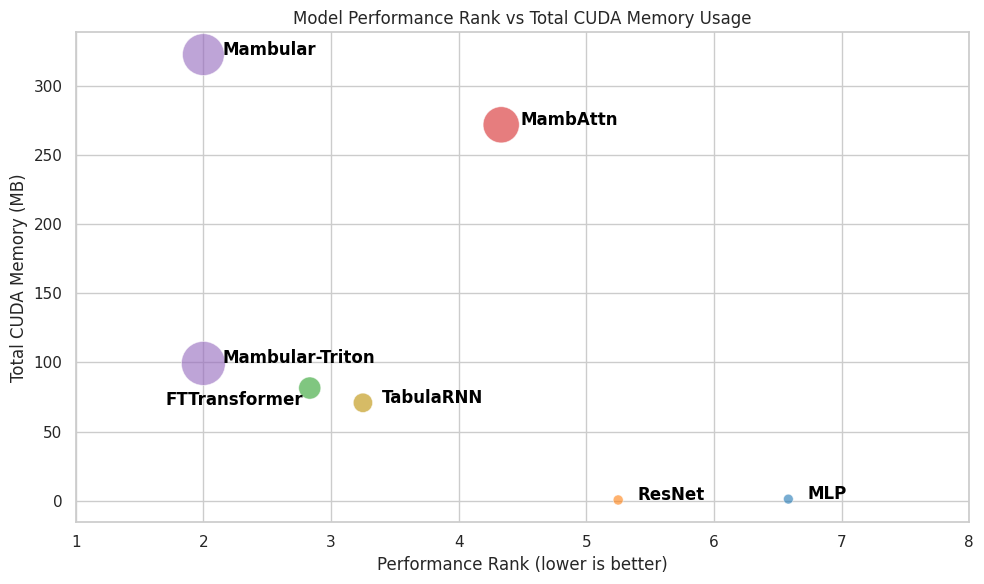}
        \caption{\footnotesize{Average rank performance vs GPU memory usage. Circle size represents the total computation time for a forward pass on a batch of 32 with 10 numerical and 10 categorical features and an embedding size of 64.}}
        \label{fig:memory_vs_performance_vs_time}
    \end{minipage}
\end{figure}


The assessment of the effect of embedding size on CUDA memory usage (Figure \ref{fig:embedding_dimension}) revealed that Mamba-based architectures' efficiency decreases at a more rapid pace than FT-Transformer and TabulaRNN with the increase in embedding size. 
Figures~\ref{fig:memory_vs_performance_vs_time} presents a comprehensive view of performance rank, CUDA memory usage, and computation time. It summarizes the results from real-world datasets (Table \ref{tab:performance_result}) and compares model efficiency with a fixed 20-feature configuration. While Mambular-Triton demonstrates an advantageous balance of low memory usage and high performance, it falls short in comparison to other evaluated alternatives when it comes to computation time. Notably, both FT-Transformer and TabulaRNN remain highly performant while being significantly more efficient than Mambular.

\begin{figure}[hb]
    \centering
    \includegraphics[width=0.49\linewidth]{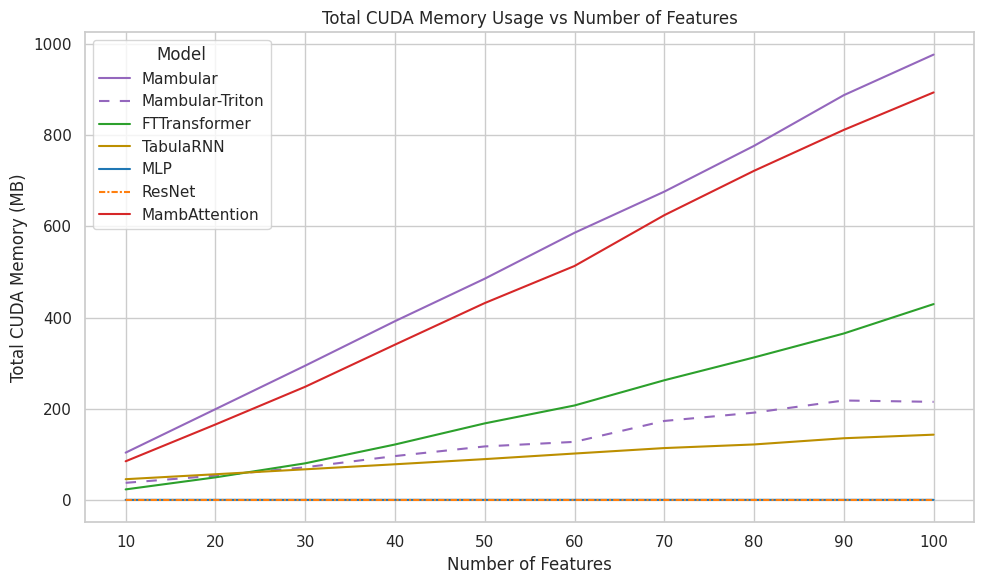}
    \includegraphics[width=0.49\linewidth]{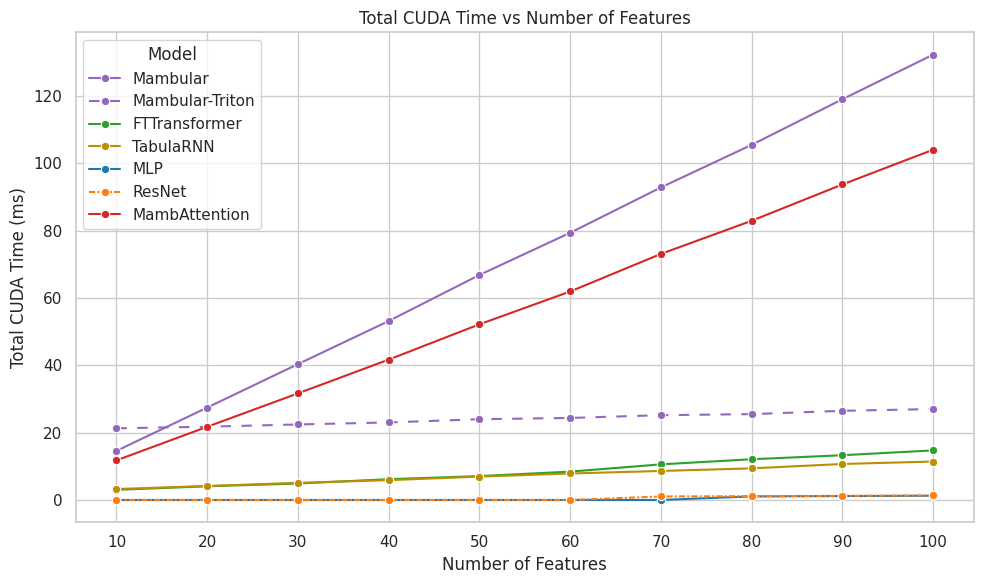}
    
        \caption{\footnotesize{Memory consumption (left panel) and GPU time taken (right panel) during a backward pass as a function of the number of features.}}
        \label{fig:backward_pass}
\end{figure}

As illustrated in Figure~\ref{fig:backward_pass}
, When analyzing training memory consumption and time, particularly during backward passes, the results closely mirror the efficiency observed during inference. Both the pure PyTorch version of Mambular and the MambAttention model exhibit linear increases in memory consumption and time taken with respect to the number of features. While the Triton version of Mambular is more efficient than the FT-Transformer for fewer than 25 features, it is slower across all tested setups. The TabulaRNN is not only more memory-efficient than the FT-Transformer, but also as fast, if not faster for larger datasets with more than 50 features.

\section{Conclusion}
With the introduction of TabulaRNN and MambAttention this study confirms that sequential models over features dimension are performant alternatives for tabular DL. It examines the transferability of efficient NLP architectures to the tabular domain, revealing that while the Mamba architecture is highly effective for text, its efficiency does not translate as well to tabular tasks, specially for the common cases in tabular domain with less than fifty features. 
Although the Mambular-Triton implementation achieves efficiency on par with FT-Transformer even for small datasets, it is slightly slower in training and inference time. 
However, the introduced TabulaRNN not only performs on par with Mambular and the attention based alternative FT-Transformer, it can outperform them in terms of efficiency and computation time. Given these results and the findings from \cite{thielmann2024mambular} regarding sequence ordering, combinations of both architectures, similar to \cite{lieber2024jamba} could be promising alternatives for tabular tasks, inheriting the performance and efficiency of TabulaRNN and the positional invariance of Transformers.

\begin{ack}
We sincerely acknowledge and appreciate the financial support provided by the Key Digital Capability (KDC) for Generative AI at BASF and the BASF Data \& AI Academy, which played a critical role in facilitating this research.
    
\end{ack}


\bibliographystyle{apalike}
\bibliography{bib}

\newpage
\appendix

\section{Datasets}\label{app:datasets}
All used datasets are taken from the UCI Machine Learning repository and publicly available\footnote{Available at https://archive.ics.uci.edu/datasets}. We drop out all missing values. For the regression tasks the targets are normalized in standard format. Note, that before PLE encoding the numerical features are scaled to be within (-1, +1) range.

\begin{table}[ht]
    \centering
    \caption{The used datasets for benchmarking. All datasets are taken from the UCI Machine Learning repository. \#num and \#cat represent the number of numerical and categorical features, respectively. The number of features thus determines the "sequence length" for Mambular and MambAttention models. The train, test and val numbers represent the average number of samples in the respective split for the 5-fold cross validation. Ratio marks the percentage of the dominant class for the binary classification tasks.}
    \begin{tabular}{l|ccccccc}
    \toprule
        Name & Abbr. & \#cat & \#num & train & test & val & ratio\\
        \hline
        & \multicolumn{7}{c}{Regression Datasets}\\
        \hline
         Diamonds           & DI  & 4	&7	&34522	&10788	&8630 &-\\
         Abalone            & AB  & 1	&8	&2673	&835	&668&-\\
         California Housing & CA  & 1	&9	&13210	&4128	&3302&-\\
         Wine Quality       & WI  & 0	&12	&4158	&1299	&1039&-\\
         Parkinsons         & PA  & 2	&20	&3760	&1175	&940&-\\
         House Sales        & HS  & 8	&19	&13832	&4322	&3458&-\\
         CPU small          & CPU & 0	&13	&5243	&1638	&1310 &-\\
        \midrule
        & \multicolumn{7}{c}{Classification Datasets}\\
        \hline
        Bank        & BA & 13	&8	&28935	&9042	&7233 & 88.3\%\\
        Adult       & AD & 9	&6	&31259	&9768	&7814 & 76.1\%\\
        Churn       & CH & 3	&9	&6400	&2000	&1600 & 79.6\%\\
        FICO        & FI & 0	&32	&6694	&2091	&1673 &53.3\%\\
        Marketing   & MA & 15	&8	&27644	&8638	&6910 &88.4\%\\	
        \bottomrule
    \end{tabular}
    
    \label{tab:datasets}
\end{table}


    

\end{document}